\documentclass[journal]{IEEEtran}

\IEEEoverridecommandlockouts

\usepackage[T1]{fontenc}
\usepackage[utf8]{inputenc}

\usepackage{newtxtext}
\usepackage{newtxmath}

\usepackage{amsmath,amssymb,amsfonts,mathtools}
\usepackage{bm}

\usepackage{booktabs,multirow}
\usepackage{tabularx}
\usepackage{array}
\usepackage{graphicx}
\graphicspath{{./}{./Figures/}{./figs/}{./Pictures/}}
\usepackage[caption=false,font=footnotesize]{subfig}

\usepackage{enumitem}
\usepackage{xcolor}
\usepackage{siunitx}
\usepackage{cite}
\usepackage[hyphens]{url}
\usepackage[hidelinks,breaklinks=true]{hyperref}
\usepackage[nameinlink]{cleveref}

\newif\ifanonymous
\anonymousfalse

\newcommand{\xr}{x_r}
\newcommand{\xs}{x_s}
\newcommand{\xri}{x_r^{(i)}}
\newcommand{\xsi}{x_s^{(i)}}
\newcommand{\SUT}{F}
\newcommand{\SUTj}{F_j}
\newcommand{\Halg}{H}

\newcommand{\Mmap}{M}
\newcommand{\Smap}{S}
\newcommand{\cvec}{\bm{c}}
\newcommand{\deltac}{\Delta\bm{c}}
\newcommand{\CDFF}{\mathrm{C\text{-}DFF}}
\newcommand{\DFF}{\mathrm{DFF}}
\newcommand{\R}{\mathbb{R}}
\newcommand{\1}{\mathbb{I}}

\newcommand{\Dcal}{\mathcal{D}}

\newcommand{\Loss}{\mathcal{L}}

\newcommand{\Tv}{T_{\mathrm{sut}}}

\newcommand{\Align}{\mathrm{Align}}
\newcommand{\Dist}{D}

\newcommand{\Distc}{D_c}
\newcommand{\Din}{D_{\mathrm{in}}}
\newcommand{\Dout}{D_{\mathrm{out}}}

\newcommand{\OV}{\mathrm{OV}}

\newcommand{\TVreg}{\mathrm{TV}}

\newcommand{\nativeRes}{\ensuremath{1248{\times}376}}

\DeclareMathOperator*{\argmin}{arg\,min}
\DeclareMathOperator*{\argmax}{arg\,max}

\title{Mechanism Parity for Simulation-Based Validation of Autonomous-Driving Perception}

\author{%

\ifanonymous
Anonymous Authors%
\else
Danial Safaei, Siddartha Khastgir, Matthew D. Higgins,~\IEEEmembership{Senior Member,~IEEE,} Jeroen Ploeg, Mohsen Alirezaei, Son Tong,~\IEEEmembership{Member,~IEEE,} Paul A. Jennings, and Xingyu Zhao%

\thanks{This work has been submitted to the IEEE for possible publication. Copyright may be transferred without notice, after which this version may no longer be accessible.}%
\thanks{This work was supported by an EPSRC-funded studentship and Siemens Industry Software Netherlands BV, administered through WMG, University of Warwick. \emph{(Corresponding author: Danial Safaei.)}}%
\thanks{This article is an extended version of the authors' conference papers \cite{Safaei2025DFF,Safaei2026CDFFITSC}. Beyond those, it contributes a unified DFF/C-DFF formalism, a single-source audited exact-geometry benchmark on ten retrained generators and three frozen SUTs, a constructed positive control with a known planted mechanism, held-out-scene replication, and multi-baseline decoupling and statistical-hardening analyses; \Cref{tab:delineation} delineates the new material.}%
\thanks{Danial Safaei, Siddartha Khastgir, Matthew D. Higgins, Paul A. Jennings, and Xingyu Zhao are with WMG, University of Warwick, Coventry, U.K. (e-mail: danial.safaei@warwick.ac.uk; s.khastgir.1@warwick.ac.uk; m.higgins@warwick.ac.uk; paul.jennings@warwick.ac.uk; Xingyu.Zhao@warwick.ac.uk).}%
\thanks{Jeroen Ploeg is with 2getthere, Utrecht, The Netherlands (e-mail: jeroen@2getthere.eu).}%
\thanks{Mohsen Alirezaei is with Eindhoven University of Technology (TU/e), Eindhoven, The Netherlands (e-mail: m.alirezaei@tue.nl).}%
\thanks{Son Tong is with Siemens Digital Industries Software, Leuven, Belgium (e-mail: son.tong@siemens.com).}%

\fi
}

\begin{document}
\maketitle

\begin{abstract}
Simulation is central to intelligent-vehicle validation, but visual realism or output agreement between a real scene and its synthetic counterpart does not prove that a perception system used the same evidence in both domains.
This paper addresses that mechanism-parity gap for paired real--synthetic autonomous-driving perception.
It unifies Decisive-Feature Fidelity, which compares decisive evidence maps in a frozen system under test (SUT), with Concept-Decisive Feature Fidelity, which projects decisive evidence into semantic scene concepts for diagnosis.
The framework is a validation diagnostic rather than a standalone safety certificate: it asks whether output-consistent simulation also preserves the mechanism by which the perception model reaches that output.
The empirical evidence comprises ten retrained diffusion ControlNet generators evaluated on a deterministic exact-geometry KITTI--Virtual KITTI~2 subset.
Generated counterparts use a 1248-by-376-pixel padded inference canvas and are cropped back to the exact KITTI dimensions before evaluation.
The final benchmark reports target-image, output-consistency, and mechanism metrics recomputed from a single audited output set, then adds metric-disagreement cases, bootstrap uncertainty, threshold sensitivity, and concept/cache integrity audits.
Across the benchmark, output and appearance fidelity are at most weakly and inconsistently rank-correlated with decisive-mechanism distance for all three frozen SUTs, and a constructed positive control shows the estimator detects a known planted mechanism that output agreement misses, supporting decisive-evidence diagnostics for intelligent-vehicle simulation validation.
All numerical claims are restricted to the declared exact-geometry benchmark, ten generators, and three frozen SUTs.
\end{abstract}


\section{Introduction}
\label{sec:intro}
\IEEEPARstart{S}{imulation} has become a core pillar of autonomous-driving development, testing, and safety assurance because it enables scalable experimentation over rare, risky, and difficult-to-reproduce traffic situations \cite{Riedmaier2020Survey,Zhong2021ScenarioSurvey}.
However, a long-standing question remains unresolved in practical validation workflows:
when a real scene and its synthetic counterpart produce similar outputs, did the system under test rely on the same evidence to make that decision?

This question matters because output agreement alone can mask a failure mode that is especially problematic for safety-critical perception:
the system can produce a correct or near-correct output for the wrong reason.
For example, a steering model may retain the same steering angle across a paired real and synthetic scene while redistributing its reliance from lane boundaries and road texture toward spurious sky artifacts or background structures.
Likewise, a segmentation model may preserve mask overlap while shifting its decisiveness from road-aligned evidence toward scene content that does not reliably transfer across domains.
Such cases weaken the validity of simulation-based conclusions even when traditional fidelity indicators appear satisfactory \cite{Lapuschkin2019CleverHans,Geirhos2020Shortcut}.

This paper builds on and unifies two complementary lines of work on that mechanism-parity problem.
The first \cite{Safaei2025DFF} introduced \emph{Decisive-Feature Fidelity} (DFF), a system-specific metric that compares decisive evidence maps extracted from a frozen system under test, and further proposed a DFF-guided generator-calibration objective.
The second \cite{Safaei2026CDFFITSC} introduced \emph{Concept-Decisive Feature Fidelity} (C-DFF), a concept-space diagnostic that aggregates decisive evidence over semantic scene concepts and identifies \emph{which} concepts become more or less decisive in synthetic imagery.
Taken together, these two lines of work suggest that simulation fidelity should be assessed not only at the level of raw inputs, latent activations, and outputs, but also at the level of decisive evidence and semantic decisiveness.

This paper unifies, extends, and hardens those ideas into a single mechanism-parity framework for simulation-based validation of autonomous-driving perception; \Cref{sec:contrib} states its specific contributions.%

The central claim of the paper is simple:
\emph{output similarity is not enough}.
What simulation-based validation requires is evidence that a frozen SUT processes matched real and synthetic scenes through sufficiently similar \emph{decisive mechanisms}, and, when this fails, diagnostics that explain where the mismatch comes from in a way that can support debugging, calibration studies, and requirement-style monitoring.

\subsection{Extension over the prior decisive-feature work}
\label{sec:journal_extension}
This article reorganizes DFF and C-DFF around one intelligent-vehicle validation question: whether synthetic counterparts preserve the decisive evidence used by frozen vehicle-perception SUTs.
\Cref{tab:delineation} delineates what is inherited from the prior conference-track work \cite{Safaei2025DFF,Safaei2026CDFFITSC} from what is introduced for the first time here, so the contribution is not mistaken for a re-presentation of the prior metrics.

\begin{table}[t]
\centering
\caption{Delineation of this paper from the prior conference-track work \cite{Safaei2025DFF,Safaei2026CDFFITSC}.}
\label{tab:delineation}
\small
\setlength{\tabcolsep}{4pt}
\renewcommand{\arraystretch}{1.15}
\begin{tabular}{@{}ll@{}}
\toprule
Component & Status \\
\midrule
DFF decisive-map metric & IV \cite{Safaei2025DFF} \\
DFF-guided generator calibration & IV \cite{Safaei2025DFF}; \emph{null} here \\
C-DFF concept-space diagnostic & ITSC \cite{Safaei2026CDFFITSC}; extended \\
Unified DFF/C-DFF formalism & new (this article) \\
Audited checkpoint regeneration & new (this article) \\
Screened qualitative audit & new (this article) \\
Disagreement \& uncertainty audits & new (this article) \\
Positive control \& sanity checks & new (this article) \\
Held-out-scene replication & new; within-corpus \\
\bottomrule
\end{tabular}
\end{table}

\subsection{Contributions}
\label{sec:contrib}
The main contributions of this paper are as follows.
\begin{itemize}[leftmargin=*,noitemsep,topsep=2pt]
 \item \textbf{Unified mechanism-parity framework.}
 We integrate DFF and C-DFF into a common formal framework for paired real--synthetic evaluation of frozen autonomous-driving perception systems.

 \item \textbf{Decisive-feature mechanism metric.}
 We formalize DFF as a system-specific measure of whether matched real and synthetic inputs induce similar decisive evidence in the SUT, including estimator choices, threshold definitions, and implementation details.

 \item \textbf{Concept-level semantic diagnostic.}
 We formalize C-DFF as a concept-space diagnostic that aggregates decisive evidence over a shared semantic taxonomy and yields interpretable concept-shift summaries.

 \item \textbf{Calibration as an evaluated optional intervention (null on this benchmark).}
 We formulate a DFF-guided calibration objective that keeps the SUT frozen and uses decisive-feature mismatch as a feedback signal for synthetic-data generation, and we treat this loop as an optional layer rather than a headline result. We report the audited checkpoints it produces and state all outcomes explicitly, including where the paired comparison shows no advantage.

 \item \textbf{Audited benchmark on retrained models.}
 We generate all benchmark evidence from the ten retrained ControlNet generators and consolidate the run manifest, generated records, aggregate metrics, per-image metrics, and figures in one auditable directory.

 \item \textbf{Diagnostic hardening.}
 We add metric-disagreement, bootstrap-uncertainty, threshold-sensitivity, and concept/cache-audit outputs that test the diagnostic value and fragility of mechanism-parity evidence under the narrow final benchmark.

 \item \textbf{Held-out-scene replication.}
 We re-evaluate all ten audited checkpoints on Virtual KITTI scenes \texttt{18}/\texttt{20}, held out from generator training, and show that the target-image parity ordering and the output/mechanism non-redundancy reproduce there---a within-corpus generalization check on two held-out scenes, not an independent-domain or corpus-wide claim.

 \item \textbf{Engineering guidance.}
 We distinguish framework claims from regenerated checkpoint evidence, report DFF/C-DFF only where recalculated from the final output cache, and state the exact empirical boundary of those claims.
\end{itemize}

\section{Related Work}
\label{sec:related}
Simulation-based validation and synthetic-data generation for automated vehicles span scenario-based safety assessment \cite{Riedmaier2020Survey,Zhong2021ScenarioSurvey}, domain randomization \cite{Tobin2017DomainRand}, high-fidelity simulators \cite{Dosovitskiy2017CARLA,Shah2018AirSim}, synthetic urban datasets \cite{Richter2016GTA,Liao2022KITTI360,Sakaridis2018FoggyCityscapes}, image translation \cite{Shrivastava2017SimGAN,Zhu2017CycleGAN,Hoffman2018CyCADA}, controllable diffusion models \cite{ControlNet2023,Rombach2022LDM}, and, recently, generative world models and neural-rendering simulators for closed-loop driving-scene synthesis \cite{Gao2024MagicDrive,Guan2024WorldModels,Yang2023UniSim,Tonderski2024NeuRAD}.
These methods can improve appearance, geometry, or domain coverage, but visual realism and target-image similarity do not prove that a particular vehicle-perception SUT relies on the same evidence in real and synthetic scenes.
Empirical studies confirm the residual gap: offline and online testing disagree \cite{Haq2021Offline}, virtual-to-physical transferability is limited \cite{Stocco2023MindTheGap}, image-translation quality metrics for the reality gap do not consistently track perception behavior \cite{Lambertenghi2024RealityGap}, the sim-to-real gap remains open across the simulation stack \cite{Hu2024SimSurvey}, and statistical treatments of scenario-based testing presuppose valid simulation evidence \cite{Zhao2025Statistical}---all quantify \emph{whether} outputs transfer, not \emph{which} evidence a fixed SUT uses, the mechanism-level question addressed here.

SUT-specific fidelity addresses that gap more directly through output-value similarity, latent-feature similarity, and safety-aware fidelity \cite{Cheng2024SAFidelity}.
Those metrics are useful, but they can still miss ``right output for the wrong reason'' cases \cite{Ross2017RRR} because they do not isolate the decisive evidence responsible for the output.
DFF targets this missing mechanism layer \cite{Safaei2025DFF}; C-DFF adds a semantic concept projection so that mechanism mismatch is interpretable in vehicle-scene terms rather than only as a heatmap distance.

The proposed metrics build on explanation methods that are well established in the interpretability literature: saliency \cite{Selvaraju2017GradCAM,Sundararajan2017IG,Smilkov2017SmoothGrad}, perturbation \cite{FongVedaldi2017MeaningfulPerturbation,Petsiuk2018RISE,Ribeiro2016LIME,Lundberg2017SHAP}, counterfactual \cite{Wachter2017Counterfactual,Goyal2019Counterfactual}, and concept-based \cite{Kim2018TCAV,Koh2020CBM} approaches, together with their faithfulness and sanity analyses \cite{Adebayo2018Sanity,Hooker2019ROAR}.

Three neighboring research lines make related mechanism-level claims but differ from the diagnostic posed here. Attribution-stability work shows that feature attributions are fragile under input perturbation \cite{Ghorbani2019Fragile} and that the distribution of a model's explanations shifts under covariate shift \cite{Mougan2023ExplanationShift}; we instead use a paired real--synthetic design with a frozen SUT to ask specifically whether the \emph{decisive} evidence is preserved, rather than monitoring explanation drift for out-of-distribution detection. Causal and interventional auditing targets shortcut learning and spurious cues---the general phenomenon DFF is meant to surface---through interventions on concepts or environments \cite{Geirhos2020Shortcut,Wu2023DISC}; DFF instead compares counterfactual decisive maps of the same fixed SUT across matched domains, yielding a simulation-validation metric rather than a training-time mitigation. Concept-based domain adaptation and concept-drift detection align or track semantic representations across domains \cite{Xu2025CUDA,Greco2024DriftLens}; C-DFF shares the concept vocabulary but \emph{tests} concept-level mechanism parity as a validation signal instead of adapting the model to close the gap. Across all three, the distinguishing elements of DFF/C-DFF are the paired real--synthetic evaluation, the frozen SUT, and the counterfactual decisive-evidence formulation used diagnostically.

\section{Problem Setting and Notation}
\label{sec:problem}
We assume paired real--synthetic image tuples $\Dcal=\{(\xri,\xsi)\}_{i=1}^{N}$, where $\xri$ is a real vehicle-scene image and $\xsi$ is its synthetic counterpart under a tightly matched scenario, camera, and frame index; the camera viewpoint matches by construction because Virtual KITTI~2 clone sequences replicate the KITTI camera trajectory frame-by-frame.
For each frozen SUT $\SUTj:\R^{H_{\mathrm{sut}}\times W_{\mathrm{sut}}\times C}\rightarrow\R^{d_j}$---where $H_{\mathrm{sut}},W_{\mathrm{sut}}$ are the SUT input height and width after preprocessing, $C$ is the number of input channels (3 for RGB), and $d_j$ is the output dimensionality of SUT $j$---all weights, preprocessing, and inference settings are fixed, since simulation fidelity is evaluated here for a fixed perception system rather than for domain adaptation.
We evaluate three frozen SUTs throughout: a steering-angle regressor (\emph{steering}), the YOLOP (You Only Look Once for Panoptic driving perception) drivable-area segmenter (\emph{YOLOP--DA}), and the YOLOP lane-line segmenter (\emph{YOLOP--LL}).
Each raw image is mapped to SUT geometry by $\Tv$, and any semantic scaffold used for C-DFF is warped into that same geometry before pooling.
Distances $\Din$, $\Dout$, $\Dist$, and $\Distc$ denote input, output, decisive-map, and concept-space discrepancies; all thresholds are descriptive operating points for the declared benchmark rather than universal guarantees.
Throughout, $S$ denotes the external semantic label map, distinct from the decisive map $M$; \Cref{tab:notation} summarizes the notation.

\begin{table}[t]
\centering
\caption{Notation and key symbols.}
\label{tab:notation}
\small
\setlength{\tabcolsep}{4pt}
\renewcommand{\arraystretch}{1.15}
\begin{tabular}{@{}ll@{}}
\toprule
Symbol & Meaning \\
\midrule
$\Dcal$ & paired benchmark $\{(\xri,\xsi)\}_{i=1}^{N}$ \\
$\xr,\ \xs$ & matched real / synthetic image \\
$\SUT$ & frozen system under test (perception model) \\
$\Tv(x)$ & mapping of a raw image to SUT input geometry \\
$\Mmap(x)$ & decisive-evidence (counterfactual) map \\
$\Smap(x)$ & external semantic label map (segmenter scaffold) \\
$\Din,\ \Dout$ & input- / output-space discrepancy \\
$\Dist$ & decisive-map distance (MSE on pooled maps) \\
$\Distc$ & concept-space distance (MSE on $\cvec$) \\
$c_k,\ \cvec$ & area-normalized concept importance / vector \\
$\DFF,\ \CDFF$ & decisive-feature / concept-decisive fidelity \\
\bottomrule
\end{tabular}
\end{table}

\section{Unified Framework Overview}
\label{sec:framework_overview}
\begin{figure*}[t]
\centering
\includegraphics[width=\textwidth]{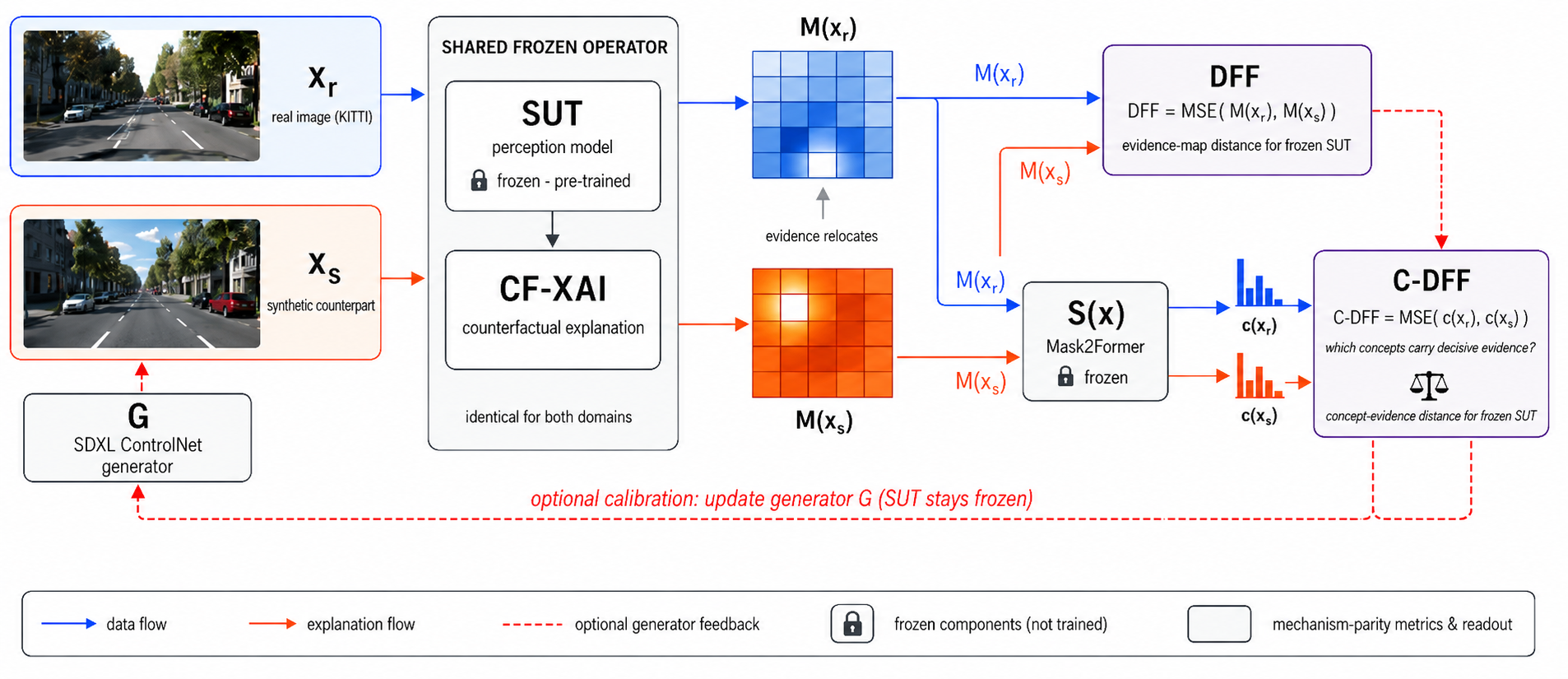}
\caption{Unified mechanism-parity framework. A matched real--synthetic pair is processed by a frozen (pre-trained) SUT and an explanation module to produce decisive maps. These are compared directly via DFF, or projected through semantic segmentation and aggregation to produce concept-importance vectors compared via C-DFF. An optional calibration loop adjusts the generator using mechanism-parity feedback---the DFF distance $\Dist(\Mmap(\xr),\Mmap(\xs))$ and, in concept-guided variants, the C-DFF distance $\Distc(\cvec(\xr),\cvec(\xs))$, each computed from \emph{both} the real and synthetic sides---while keeping the SUT fixed.}
\label{fig:journal_overview}
\end{figure*}

Before formalizing DFF and C-DFF separately, it is helpful to view them as two levels of the same mechanism-parity framework, summarized in \Cref{fig:journal_overview}.

For a matched pair $(\xr,\xs)$ and frozen SUT $\SUT$, the workflow has three conceptual layers:
\begin{enumerate}[leftmargin=*,noitemsep,topsep=2pt]
 \item \textbf{Direct decisive-evidence comparison.}
 Extract decisive evidence maps $\Mmap(\xr)$ and $\Mmap(\xs)$ and compare them directly.
 This yields DFF.

 \item \textbf{Semantic projection and aggregation.}
 Project decisive evidence into a concept space using an external semantic segmentation scaffold, aggregate concept-level importance, and compare the resulting vectors.
 This yields C-DFF.

 \item \textbf{Optional synthetic-data calibration.}
 Use DFF as a mechanism-parity feedback signal inside a calibration loop that adjusts generator parameters while keeping the SUT fixed.
\end{enumerate}

\section{Decisive-Feature Fidelity}
\label{sec:dff}

Input fidelity, output fidelity, latent-feature fidelity, and safety-aware fidelity are useful but incomplete for mechanism parity: high visual or output agreement can still be a right-output-for-the-wrong-reason failure.
We instantiate the decisive map with counterfactual-style evidence, treating it as a diagnostic approximation of intervention-style decisiveness---how strongly the output would change if that evidence were altered---rather than proof of a full structural causal model \cite{Pearl2009Causality}; because the explanation module is noisy and hyperparameter-sensitive, we keep the SUT frozen, report mechanism metrics only inside the declared paired benchmark, and make the approximation limits explicit.
For a fixed SUT $\SUT$ and a generic input image $x$ (either the real image $\xr$ or its synthetic counterpart $\xs$), let
\begin{equation}
\Mmap(x) = \Halg(\SUT, \Tv(x))
\in \R_{\ge 0}^{H_{\mathrm{sut}}\times W_{\mathrm{sut}}},
\label{eq:dmap_general}
\end{equation}
where the explanation operator $\Halg$ maps a frozen SUT and an image in SUT geometry to a nonnegative decisive map,
$\Halg:\mathcal{F}\times[0,1]^{H_{\mathrm{sut}}\times W_{\mathrm{sut}}\times C}\rightarrow\R_{\ge 0}^{H_{\mathrm{sut}}\times W_{\mathrm{sut}}}$, with $\mathcal{F}$ the set of frozen SUTs; \Cref{sec:dff_estimator} defines $\Halg$ explicitly through \eqref{eq:cf_obj}--\eqref{eq:cf_select}.

\subsection{DFF definition}
\label{sec:dff_definition}
Given a matched pair $(\xr,\xs)$ and frozen SUT $\SUT$, we define Decisive-Feature Fidelity by the decisive-map distance $\Dist$:
\begin{equation}
\DFF(\xr,\xs;\SUT)
\coloneqq
\Dist\!\big(\Mmap(\xr),\Mmap(\xs)\big).
\label{eq:dff_def}
\end{equation}
Lower DFF indicates stronger mechanism parity at the decisive-feature level.

\subsection{Estimator instantiation}
\label{sec:dff_estimator}
We instantiate $\Halg$ by optimizing a counterfactual image $x_{\mathrm{cf}}$, initialized at the SUT-geometry input $x_{\mathrm{cf}}^{(0)}=\Tv(x)$, that drives the SUT's scalar task summary toward a target while staying close to the original input. Concretely, we \emph{minimize}
\begin{equation}
\mathcal{J}_t(x_{\mathrm{cf}})
=
\bigl(s(x_{\mathrm{cf}})-s_{\mathrm{tgt}}\bigr)^2
+
w_t\,\mathrm{mean}\!\bigl(|x_{\mathrm{cf}}-\Tv(x)|\bigr)
+
\lambda_{\mathrm{TV}}\,\TVreg(x_{\mathrm{cf}}),
\label{eq:cf_obj}
\end{equation}
where $s(\cdot)$ is a task-dependent scalar summary of the SUT output, $s_{\mathrm{tgt}}$ is its target-side value, $w_t$ is a proximity weight at optimization step $t$ (schedule below), $\TVreg(\cdot)$ is the total-variation regularizer \cite{Rudin1992TV} with weight $\lambda_{\mathrm{TV}}$, and $|\cdot|$ is element-wise. Optimizing $\mathcal{J}_t$ by gradient descent yields a trajectory of iterates, from which the returned counterfactual $x_{\mathrm{cf}}^{\star}$ is selected by \eqref{eq:cf_select} rather than by the running objective value; the decisive map of \eqref{eq:dmap_general} is its channel-mean absolute perturbation,
\begin{equation}
\Mmap(x) = \tfrac{1}{C}\textstyle\sum_{c=1}^{C}\bigl|x_{\mathrm{cf}}^{\star}-\Tv(x)\bigr|_{c}.
\label{eq:dmap_from_cf}
\end{equation}
Counterfactuals are optimized with Adam over the input pixels: the steering SUT uses $80$ steps at learning rate $3\times10^{-2}$, whereas the YOLOP drivable-area and lane-line SUTs use $60$ steps at $2\times10^{-2}$; all three share $\lambda_{\mathrm{TV}}=10^{-3}$ and a two-stage proximity weight $w_t$ that rises from $0.1$ in the first half of the schedule to $0.5$ (steering) or $0.3$ (YOLOP) thereafter.
In the released implementation, $s(\cdot)$ is the mean predicted steering value (steering) or the mean foreground probability of the corresponding mask head (YOLOP), and $s_{\mathrm{tgt}}=s(\Tv(x))+\delta$ offsets the image's own prediction by $\delta$ with $|\delta|=0.1$, away from its current sign or decision side; every iterate is projected onto the valid image range $[0,1]$. Let $\mathcal{A}=\{t:\lvert s(x_{\mathrm{cf}}^{(t)})-s(\Tv(x))\rvert\ge 0.8\,\lvert\delta\rvert\}$ collect the iterates reaching at least $80\%$ of the target change. The returned counterfactual is then the minimum-perturbation accepted iterate,
\begin{equation}
x_{\mathrm{cf}}^{\star}=\argmin_{t\in\mathcal{A}}\ \mathrm{mean}\bigl(\lvert x_{\mathrm{cf}}^{(t)}-\Tv(x)\rvert\bigr),
\label{eq:cf_select}
\end{equation}
falling back to $\argmax_{t}\lvert s(x_{\mathrm{cf}}^{(t)})-s(\Tv(x))\rvert$ when $\mathcal{A}=\emptyset$; composed with \eqref{eq:dmap_from_cf}, this defines $\Halg$.
Because step counts, learning rates, and the scalar summary $s(\cdot)$ differ across SUTs, DFF magnitudes are calibrated per SUT and are not directly comparable \emph{between} SUTs; we therefore compare DFF only within a fixed SUT, while its qualitative interpretation (which evidence shifts) carries across SUTs.
For the reported metrics, $\Dist$ is the MSE on the pooled decisive maps:
\begin{equation}
\Dist\!\big(\Mmap(\xr),\Mmap(\xs)\big)
=
\frac{1}{hw}
\sum_{p=1}^{hw}
\left(
\widetilde{\Mmap}(\xr)_p - \widetilde{\Mmap}(\xs)_p
\right)^2,
\label{eq:dff_distance_mse}
\end{equation}
where $\widetilde{\Mmap}$ denotes average pooling to a fixed $h\times w$ grid.
DFF therefore measures decisive-evidence similarity rather than raw appearance, broad hidden representations, or final outputs; divergence from image or output metrics is the intended mechanism-shift signal.

\section{Concept-Decisive Feature Fidelity}
\label{sec:cdff}
C-DFF adds semantic actionability by projecting decisive evidence into a shared urban-scene vocabulary.
Let $K$ be the number of semantic classes in the external segmenter, and let
\begin{equation}
\Smap(x)\in \{1,\dots,K\}^{H_{\mathrm{sut}}\times W_{\mathrm{sut}}}
\end{equation}
be its off-the-shelf semantic label map, assigning each pixel a class index in $\{1,\dots,K\}$; the Mask2Former/Cityscapes segmenter is frozen and used only as a scaffold \cite{Cheng2022Mask2Former,Cordts2016Cityscapes,HFMask2FormerCityscapes}. After nearest-neighbor warping into SUT geometry---$\Smap_{\mathrm{sut}}(x)$ denotes $\Smap(x)$ warped into that geometry---decisive evidence and semantic labels are pooled onto a common grid:
\begin{equation}
\bigl(\widetilde{\Mmap}(x), \widetilde{\Smap}(x)\bigr)
=
\Align\!\bigl(\Mmap(x), \Smap_{\mathrm{sut}}(x)\bigr),
\label{eq:align}
\end{equation}
with primary choice $h=w=16$.

Let $\Omega=\{(i,j):1\le i\le h,\ 1\le j\le w\}$ index the pooled grid, and let $\1[\cdot]$ be the indicator function (one if its argument holds, zero otherwise). For concept $k\in\{1,\dots,K\}$ define
\begin{align}
A_k(x)
&=
\sum_{(i,j)\in\Omega}
\1[\widetilde{\Smap}(x)_{i,j} = k],\\
B_k(x)
&=
\sum_{(i,j)\in\Omega}
\1[\widetilde{\Smap}(x)_{i,j} = k]\,\widetilde{\Mmap}(x)_{i,j}.
\end{align}
The area-normalized concept importance is
\begin{equation}
c_k(x)
=
\frac{B_k(x)}{\max(1,A_k(x))}.
\label{eq:concept_importance}
\end{equation}
Stacking these values gives the concept-importance vector
\begin{equation}
\cvec(x)
=
[c_1(x),\dots,c_K(x)]^\top
\in \R_{\ge 0}^{K}.
\label{eq:cvec}
\end{equation}

Given a matched pair $(\xr,\xs)$ and frozen SUT $\SUT$, define
\begin{equation}
\CDFF(\xr,\xs;\SUT)
=
\Distc\!\bigl(\cvec(\xr), \cvec(\xs)\bigr).
\label{eq:cdff_def}
\end{equation}
Our primary distance is concept-space MSE:
\begin{equation}
\Distc\bigl(\cvec(\xr),\cvec(\xs)\bigr)
=
\frac{1}{K}\sum_{k=1}^{K}\bigl(c_k(\xr)-c_k(\xs)\bigr)^2.
\label{eq:cdff_mse}
\end{equation}
Lower C-DFF indicates stronger mechanism parity in semantic concept space.
A scalar C-DFF indicates mismatch magnitude but not cause, so we also define the per-pair concept-shift vector
\begin{equation}
\deltac(\xr,\xs)
=
\cvec(\xs)-\cvec(\xr).
\label{eq:deltac}
\end{equation}
Interpretation is immediate:
$\Delta c_k>0$ means concept $k$ becomes more decisive in synthetic imagery, while $\Delta c_k<0$ means it becomes less decisive.
DFF answers whether decisive mechanisms match; C-DFF explains which concepts account for mismatch.

\section{DFF-Guided Calibration}
\label{sec:calibration}
DFF can also be used as an optional intervention signal. In the implemented calibration variants, ControlNet parameters are updated while downstream SUT heads remain frozen. For a mini-batch, the implemented loss is
\begin{equation}
\Loss
=
\Loss_{\mathrm{recon}}
+
\tilde{\beta}\,\Loss_{\OV}
+
\tilde{\lambda}\,\Loss_{\DFF}
+
\lambda_c\,\Loss_{\mathrm{concept}},
\label{eq:calibration_obj}
\end{equation}
where $\Loss_{\mathrm{recon}}$ is the diffusion (noise-prediction) reconstruction loss to the matched target, $\Loss_{\OV}=\Dout$ is the output-value discrepancy between the SUT outputs on the synthetic and real images, $\Loss_{\DFF}=\Dist(\Mmap(\xr),\Mmap(\xs))$ reuses the decisive-map distance of \eqref{eq:dff_distance_mse}, and $\Loss_{\mathrm{concept}}=\Distc(\cvec(\xr),\cvec(\xs))$ reuses the concept-space distance of \eqref{eq:cdff_mse} (enabled in concept-guided variants, omitted in DFF-only variants); $\tilde{\beta},\tilde{\lambda},\lambda_c\ge 0$ are tuned base loss weights.
In the released implementation these base weights are applied as mild data-dependent multipliers (scaled by per-batch conditioning complexity and road content), and in the concept-guided variants $\Loss_{\mathrm{concept}}$ additionally penalizes a fixed set of undesired scene concepts (sky, vegetation, building, and terrain, weighted equally to the concept-alignment term); the exact schedules and settings are preserved in the reproduction package. Because this penalty targets the same large-area concepts examined by the concept-shift analysis, the concept-guided variants' reported concept shifts partly reflect the calibration objective.
The final training runs use reduced differentiable counterfactual loops for tractability---a short optimization horizon and a larger, sign-flipping output target unlike the evaluation's $\pm 0.1$ offset---while final reported DFF/C-DFF mechanism metrics are recomputed offline from exact-size outputs with the evaluation counterfactual-XAI (CF-XAI) configuration, so the reported mechanism metrics do not depend on the weaker training-time counterfactual signal.
Outer-loop tuning uses the Optuna Tree-structured Parzen Estimator (TPE) \cite{Akiba2019Optuna}; exact generated scripts and settings are preserved in the reproduction package.
The calibrated variants are evaluated against the baseline generator within the benchmark of \Cref{sec:final_retrain_benchmark}, which reports the outcome of this loop.

\section{Experimental Protocol}
\label{sec:experiments}
This section reports the mechanism-parity benchmark on paired exact-geometry KITTI--Virtual KITTI~2 scenes; the objective is to test whether DFF and C-DFF expose disagreement between output-level fidelity and mechanism-level evidence for the three frozen SUTs introduced in \Cref{sec:problem} (steering, YOLOP--DA, YOLOP--LL).
All quantitative evidence in this section is generated from the ten retrained checkpoints and recomputed from a single audited output cache; every reported target-image, output-consistency, DFF, and C-DFF value comes from that one source, with no quantity carried over from elsewhere.
The qualitative evidence combines two audited surfaces: panels from the exact-size output cache, and a fresh random-candidate failure screen that tests numerically whether steering-targeted visual failures are systematic.
The DFF/C-DFF mechanism values are recomputed from the exact-size output cache using the estimators defined in \Cref{sec:dff,sec:cdff}.

The evaluation source is the canonical target-supervised KITTI--Virtual KITTI~2 dataset, whose rows contain KITTI RGB source, Virtual KITTI RGB segmentation conditioning, Virtual KITTI RGB target, and prompt fields.
The regenerated benchmark uses 96 exact-geometry source/conditioning/target triplets sampled from 950 dimension-matched rows within 2126 valid KITTI--Virtual KITTI~2 metadata rows \cite{Geiger2013KITTI,Cabon2020VKITTI2}, using sample seed \texttt{20260505}; each triplet yields one real--synthetic evaluation pair. The 96 pairs span three Virtual KITTI scenes (54 from scene \texttt{01}, 22 from scene \texttt{02}, and 20 from scene \texttt{06}), so the larger sample increases statistical power rather than scene diversity. The selected pair identifiers are preserved in the run manifest.
This subset defines the deterministic exact-geometry benchmark used here; results are restricted to these three Virtual KITTI scenes.

The benchmark evaluates ten independently retrained SDXL-based ControlNet generators trained under different loss configurations \cite{ControlNet2023,Rombach2022LDM}: one baseline (reconstruction loss only), three output-value-fidelity-guided (OVF) variants, three DFF-guided variants, and three C-DFF (concept-guided) variants; the nine guided variants differ in the loss weights $\tilde{\beta},\tilde{\lambda},\lambda_c$ of \eqref{eq:calibration_obj}, tuned by Optuna, and all ten are frozen during evaluation.
All checkpoints were retrained at the native \nativeRes{} resolution, used target-image supervision, and disabled intermediate checkpointing.
Generation uses prompt ``virtual kitti style'', 30 denoising steps, guidance scale 7.5, ControlNet scale 0.8, base inference seed \texttt{12345} (offset per pair), and a padded-inference scheme: the image is right/bottom padded to the native \nativeRes{} canvas, run through the generator, and cropped back to the exact KITTI dimensions.
The manuscript reports target-image, output-consistency, DFF, and C-DFF mechanism metrics.
Output-consistency metrics are computed against the KITTI source using frozen steering and YOLOP-style SUTs \cite{Bojarski2016E2E,Chernikova2019Udacity,Udacity2016Challenge2,Wu2022YOLOP}: steering consistency uses $\exp(-|\theta_r-\theta_s|/0.2)$, where $\theta_r$ and $\theta_s$ are the steering values predicted on the real and synthetic images; drivable-area consistency uses mask IoU; and lane-line consistency uses a Chamfer/F1-style YOLOP lane-line score implemented in the released evaluation code \cite{Borgefors1988Chamfer}.
DFF/C-DFF metrics are then computed from the same exact-size generated outputs using the CF-XAI decisive-map estimator and Mask2Former concept aggregation defined earlier in the paper.

\section{Final Retrained-Model Audit}
\label{sec:final_retrain_audit}
The saved-model audit is mechanical and intentionally independent of visual quality.
Each final output directory must exist, include \texttt{config.json}, include \texttt{diffusion\_pytorch\_model.safetensors}, parse through \texttt{safe\_open} on CPU, and have a latest training log with return code zero.
All ten final checkpoints pass these criteria.
The row-level artifact table is released with the reproduction package; each row has tensor count 844, the expected safetensors size, and status \texttt{pass}.

\section{Final Exact-Geometry Benchmark}
\label{sec:final_retrain_benchmark}
\subsection{Aggregate quantitative results}
\Cref{tab:final_retrain_benchmark} reports the regenerated aggregate metrics. MAE, PSNR, and SSIM are standard image-similarity measures between the generated image and its matched Virtual KITTI target (mean absolute error, peak signal-to-noise ratio, structural similarity); they are reported as conventional perceptual baselines, not training objectives---generation is conditioned on task-aligned semantics, not optimized for realism---and their decoupling from the mechanism metrics is itself part of the finding. Output-consistency (OV) is the per-SUT real-vs-synthetic output agreement defined below.
On this regenerated subset, the lowest mean target MAE is achieved by Baseline (\num{0.0918}), while the highest mean target MAE is observed for DFF steering (\num{0.2881}). The highest mean SSIM is achieved by Baseline (\num{0.5946}).
The steering-targeted variants are the weakest by target-image metrics, while the YOLOP-targeted variants stay close to their matched targets.
A separate fresh random-candidate failure audit confirms that this is not just one unlucky qualitative frame: variants 02, 05, and 08 are flagged as scene-failure outliers on all 36 steering-targeted outputs of this screen (12 source frames evaluated across the three steering-targeted variants \texttt{02}/\texttt{05}/\texttt{08}, i.e.\ $12\times3=36$ outputs) using the rule SSIM $<0.40$ or MAE $>0.18$.
We therefore treat the steering-targeted checkpoints as failed/outlier checkpoints for publication-quality counterpart generation, while keeping their quantitative rows in the audit trail.
The pattern does not indicate a global SDXL ControlNet failure, because the baseline and YOLOP-targeted OVF/DFF/C-DFF variants remain visually aligned on the same source frame; it instead points to instability in the steering-targeted guidance objective/checkpoints under this final training configuration.

\begin{table*}[t]
\centering
\caption{Final benchmark on the regenerated evaluation subset. All numbers are calculated from the generated images. MAE is lower-better; PSNR, SSIM, and output-consistency scores are higher-better. Values are mean $\pm$ one standard deviation across the 96 paired images.}
\label{tab:final_retrain_benchmark}
\footnotesize
\setlength{\tabcolsep}{3pt}
\renewcommand{\arraystretch}{1.08}\resizebox{\textwidth}{!}{%
\begin{tabular}{@{}cllcccccc@{}}
\toprule
\textbf{ID} & \textbf{Generator} & \textbf{Family} & \textbf{MAE} & \textbf{PSNR} & \textbf{SSIM} & \textbf{Steer OV} & \textbf{DA OV} & \textbf{LL OV} \\
\midrule
01 & Baseline & Baseline & \num{0.0918}$\pm$\num{0.0239} & \num{16.73}$\pm$\num{2.05} & \num{0.5946}$\pm$\num{0.0915} & \num{0.3589}$\pm$\num{0.2893} & \num{0.3841}$\pm$\num{0.2336} & \num{0.1138}$\pm$\num{0.1023} \\
02 & OVF steering & OVF & \num{0.2817}$\pm$\num{0.0372} & \num{9.15}$\pm$\num{1.12} & \num{0.2453}$\pm$\num{0.0626} & \num{0.2474}$\pm$\num{0.2867} & \num{0.5143}$\pm$\num{0.2063} & \num{0.1058}$\pm$\num{0.0812} \\
03 & OVF YOLOP--DA & OVF & \num{0.1311}$\pm$\num{0.0521} & \num{14.95}$\pm$\num{2.96} & \num{0.5273}$\pm$\num{0.1224} & \num{0.3828}$\pm$\num{0.3192} & \num{0.4421}$\pm$\num{0.2178} & \num{0.1119}$\pm$\num{0.0912} \\
04 & OVF YOLOP--LL & OVF & \num{0.1050}$\pm$\num{0.0301} & \num{15.95}$\pm$\num{2.15} & \num{0.5796}$\pm$\num{0.0935} & \num{0.3691}$\pm$\num{0.3091} & \num{0.4014}$\pm$\num{0.2300} & \num{0.1159}$\pm$\num{0.1099} \\
05 & DFF steering & DFF & \num{0.2881}$\pm$\num{0.0511} & \num{8.62}$\pm$\num{1.37} & \num{0.2495}$\pm$\num{0.0609} & \num{0.2450}$\pm$\num{0.2556} & \num{0.5827}$\pm$\num{0.1779} & \num{0.1601}$\pm$\num{0.1047} \\
06 & DFF YOLOP--DA & DFF & \num{0.1077}$\pm$\num{0.0352} & \num{16.02}$\pm$\num{2.69} & \num{0.5560}$\pm$\num{0.1138} & \num{0.3408}$\pm$\num{0.3085} & \num{0.4935}$\pm$\num{0.2113} & \num{0.1288}$\pm$\num{0.1050} \\
07 & DFF YOLOP--LL & DFF & \num{0.0966}$\pm$\num{0.0252} & \num{16.46}$\pm$\num{2.00} & \num{0.5823}$\pm$\num{0.0913} & \num{0.3505}$\pm$\num{0.2968} & \num{0.4142}$\pm$\num{0.2362} & \num{0.1190}$\pm$\num{0.1198} \\
08 & C-DFF steering concept & C-DFF & \num{0.2868}$\pm$\num{0.0448} & \num{8.80}$\pm$\num{1.26} & \num{0.2557}$\pm$\num{0.0559} & \num{0.2814}$\pm$\num{0.2948} & \num{0.4301}$\pm$\num{0.2311} & \num{0.0951}$\pm$\num{0.0876} \\
09 & C-DFF YOLOP--DA concept & C-DFF & \num{0.1083}$\pm$\num{0.0376} & \num{15.94}$\pm$\num{2.69} & \num{0.5585}$\pm$\num{0.1081} & \num{0.3297}$\pm$\num{0.2788} & \num{0.4693}$\pm$\num{0.2118} & \num{0.1144}$\pm$\num{0.1043} \\
10 & C-DFF YOLOP--LL concept & C-DFF & \num{0.0941}$\pm$\num{0.0241} & \num{16.49}$\pm$\num{2.08} & \num{0.5859}$\pm$\num{0.0956} & \num{0.3367}$\pm$\num{0.2920} & \num{0.4068}$\pm$\num{0.2293} & \num{0.1187}$\pm$\num{0.1051} \\
\bottomrule
\end{tabular}%
}
\end{table*}

\subsection{Final DFF/C-DFF mechanism results}
\Cref{tab:final_mechanism_metrics} reports the regenerated mechanism metrics from the exact same final output cache used for \Cref{tab:final_retrain_benchmark}.
Per-generator differences are small relative to their per-pair dispersion (the mechanism distributions are strongly right-skewed, so we compare their dispersion rather than reading a symmetric $\pm$one-standard-deviation band), and cross-explainer concordance (\Cref{sec:threats}) makes per-generator ordering unreliable for steering and YOLOP--LL---illustratively, a visually failed generator (OVF steering) attains the lowest mean C-DFF on the YOLOP--LL SUT (\num{2.66e-05})---so we read \Cref{tab:final_mechanism_metrics} as a population-level summary rather than a per-generator ranking.
For YOLOP--DA, where the two explainers agree, the concept-guided variant attains the lowest mean DFF (\num{1.43e-04}) and C-DFF (\num{8.35e-05}).
These mechanism values are reported alongside the image and output-consistency metrics above, which can score the same generator differently.
A paired Wilcoxon signed-rank comparison of each guided variant against the baseline over the 96 pairs finds no SUT with a significant mechanism-metric improvement after Holm correction, so the calibration loop of \Cref{sec:calibration} is reported as an evaluated negative result on this subset. The point estimates are small---even the most favorable guided variant per SUT has a paired effect size on decisive-map distance of $|d_z|\le 0.11$ (best-case $d_z=+0.05$, $-0.11$, $+0.02$ for steering/YOLOP--DA/YOLOP--LL) and every bootstrap $95\%$ interval spans zero---but at $n=96$ these intervals are wide (up to ${\approx}{\pm}0.25$ in $d_z$), so we report an evaluated null at limited power rather than a formal equivalence result.
This is a genuine negative finding that runs counter to the positive calibration result of \cite{Safaei2025DFF}: there, a lightweight calibrator adjusting generator parameters on a larger 80/20 scene split improved decisive-feature fidelity, whereas here, when the ControlNet weights themselves are retrained under the stricter audited exact-geometry protocol, DFF-guided calibration shows no significant mechanism-metric advantage over the reconstruction baseline. Two design changes co-vary between the studies---the calibrator class (lightweight parameter adjustment vs.\ full-weight ControlNet retraining) and the evaluation protocol (80/20 split vs.\ exact-geometry $N{=}96$)---so we cannot attribute the null to one cause: it may reflect that full-weight retraining is a harder optimization for a mechanism-parity signal, or that the exact-geometry protocol is simply more demanding. A bridging experiment that reproduces the lightweight calibrator under the exact-geometry protocol would separate these hypotheses and is left to future work; we report the null as an evaluated outcome rather than claiming the earlier improvement transfers.

\begin{table*}[t]
\centering
\caption{Final exact-size DFF/C-DFF mechanism metrics on the regenerated benchmark outputs. DFF is pooled decisive-map MSE; C-DFF is concept-importance-vector MSE. All values are recomputed from the exact-size generated outputs after cropping the padded inference canvas back to the KITTI source dimensions, with steering CF-XAI at 80~steps and YOLOP CF-XAI at 60~steps. Lower is better; values are mean $\pm$ one standard deviation across the 96 pairs. Given the heterogeneous cross-explainer concordance (\Cref{sec:threats}), rows are population-level summaries rather than a per-generator ranking.}
\label{tab:final_mechanism_metrics}
\footnotesize
\setlength{\tabcolsep}{2.6pt}
\renewcommand{\arraystretch}{1.08}
\resizebox{\textwidth}{!}{%
\begin{tabular}{@{}cllcccccc@{}}
\toprule
\textbf{ID} & \textbf{Generator} & \textbf{Family} & \textbf{DFF-S} & \textbf{C-DFF-S} & \textbf{DFF-DA} & \textbf{C-DFF-DA} & \textbf{DFF-LL} & \textbf{C-DFF-LL} \\
\midrule
01 & Baseline & Baseline & \num{2.38e-04}$\pm$\num{2.45e-04} & \num{2.72e-04}$\pm$\num{2.69e-04} & \num{1.50e-04}$\pm$\num{1.13e-04} & \num{1.02e-04}$\pm$\num{1.01e-04} & \num{1.27e-05}$\pm$\num{4.29e-06} & \num{2.91e-05}$\pm$\num{1.75e-05} \\
02 & OVF steering & OVF & \num{3.64e-04}$\pm$\num{2.91e-04} & \num{2.88e-04}$\pm$\num{1.63e-04} & \num{1.96e-04}$\pm$\num{1.58e-04} & \num{1.30e-04}$\pm$\num{1.33e-04} & \num{1.39e-05}$\pm$\num{5.16e-06} & \num{2.66e-05}$\pm$\num{1.39e-05} \\
03 & OVF YOLOP--DA & OVF & \num{2.95e-04}$\pm$\num{2.63e-04} & \num{3.23e-04}$\pm$\num{3.56e-04} & \num{1.49e-04}$\pm$\num{1.33e-04} & \num{9.42e-05}$\pm$\num{9.08e-05} & \num{1.28e-05}$\pm$\num{5.30e-06} & \num{2.85e-05}$\pm$\num{1.70e-05} \\
04 & OVF YOLOP--LL & OVF & \num{3.10e-04}$\pm$\num{2.74e-04} & \num{2.82e-04}$\pm$\num{2.32e-04} & \num{1.49e-04}$\pm$\num{1.03e-04} & \num{1.12e-04}$\pm$\num{1.05e-04} & \num{1.33e-05}$\pm$\num{5.11e-06} & \num{3.29e-05}$\pm$\num{1.76e-05} \\
05 & DFF steering & DFF & \num{3.36e-04}$\pm$\num{2.97e-04} & \num{2.63e-04}$\pm$\num{1.66e-04} & \num{1.58e-04}$\pm$\num{1.33e-04} & \num{1.10e-04}$\pm$\num{9.60e-05} & \num{1.87e-05}$\pm$\num{1.10e-05} & \num{3.39e-05}$\pm$\num{1.63e-05} \\
06 & DFF YOLOP--DA & DFF & \num{2.68e-04}$\pm$\num{2.60e-04} & \num{3.11e-04}$\pm$\num{2.57e-04} & \num{1.56e-04}$\pm$\num{1.21e-04} & \num{9.23e-05}$\pm$\num{8.20e-05} & \num{1.28e-05}$\pm$\num{5.52e-06} & \num{2.96e-05}$\pm$\num{1.59e-05} \\
07 & DFF YOLOP--LL & DFF & \num{2.72e-04}$\pm$\num{2.51e-04} & \num{3.00e-04}$\pm$\num{2.70e-04} & \num{1.49e-04}$\pm$\num{1.22e-04} & \num{9.52e-05}$\pm$\num{7.94e-05} & \num{1.37e-05}$\pm$\num{5.55e-06} & \num{3.10e-05}$\pm$\num{1.83e-05} \\
08 & C-DFF steering concept & C-DFF & \num{3.36e-04}$\pm$\num{2.81e-04} & \num{2.61e-04}$\pm$\num{1.52e-04} & \num{1.62e-04}$\pm$\num{1.25e-04} & \num{1.03e-04}$\pm$\num{9.31e-05} & \num{1.70e-05}$\pm$\num{8.22e-06} & \num{3.16e-05}$\pm$\num{1.59e-05} \\
09 & C-DFF YOLOP--DA concept & C-DFF & \num{2.89e-04}$\pm$\num{2.49e-04} & \num{2.76e-04}$\pm$\num{2.00e-04} & \num{1.43e-04}$\pm$\num{9.78e-05} & \num{8.35e-05}$\pm$\num{7.47e-05} & \num{1.36e-05}$\pm$\num{6.78e-06} & \num{2.99e-05}$\pm$\num{1.62e-05} \\
10 & C-DFF YOLOP--LL concept & C-DFF & \num{2.48e-04}$\pm$\num{2.25e-04} & \num{2.65e-04}$\pm$\num{1.85e-04} & \num{1.57e-04}$\pm$\num{1.22e-04} & \num{9.90e-05}$\pm$\num{9.78e-05} & \num{1.34e-05}$\pm$\num{6.22e-06} & \num{3.08e-05}$\pm$\num{1.64e-05} \\
\bottomrule
\end{tabular}%
}
\end{table*}

\subsection{Diagnostic value beyond visual and output metrics}
\label{sec:tiv_diagnostic_value}
A key empirical question is whether DFF/C-DFF add information beyond target-image metrics and frozen-SUT output consistency.
To test this, we added a post-hoc hardening analysis over the same final output cache.
For each generated image and SUT, the analysis ranks apparent fidelity by combining SSIM-to-target rank with output-consistency rank, then ranks mechanism mismatch by the worse of the DFF and C-DFF ranks.
Across the 2880 SUT--variant--pair rows, the high-apparent-fidelity/high-mechanism-mismatch rule identifies 426 rows: 163 for steering, 116 for YOLOP--DA, and 147 for YOLOP--LL.
All of the top 24 ranked disagreement cases satisfy this rule, and the released CSV records representative cases such as pair \texttt{0006\_000150} under steering and pair \texttt{0002\_000109} under YOLOP--LL.
These rows are precisely the cases in which visual and output-level checks alone would be least informative: the generator can look close to the target and preserve the SUT output while still shifting the decisive evidence measured by DFF/C-DFF.
The two summary good-fractions differ markedly---mechanism good-fraction \num{0.369}/\num{0.408}/\num{0.345} versus \num{0.500} apparent for steering/YOLOP--DA/YOLOP--LL---but this gap is partly definitional and we therefore do not attach a significance test to it: the apparent good-fraction averages two rank percentiles and is close to \num{0.500} by construction, whereas the mechanism score takes the worse of the DFF and C-DFF ranks, so a paired test of one against the other measures a built-in order-statistic difference rather than an operational effect size. The substantive divergence evidence is instead the disagreement-case count above and the effect-size correlation analysis in \Cref{sec:discussion}, which use consistently defined quantities.

We also added three boundary checks around this diagnostic use.
First, bootstrap confidence intervals over the fixed 96-pair subset are reported; the widest relative interval is \num{0.468} (variant 02, steering output consistency; widest mechanism interval \num{0.438}, variant 03 C-DFF MSE on the steering SUT), so the small-sample uncertainty is shown rather than hidden.
Second, a threshold-sensitivity table reports 360 variant/SUT/metric operating-point rows using 75/85/90/95 percentile thresholds, so the paper does not rely on one hand-picked mechanism threshold.
Third, concept-shift, segmentation-distribution, and cache-integrity audits cover the C-DFF concept-projection layer and the cached decisive maps/segmentations (per cache: 96 source and 960 generated entries).
The largest mean absolute concept shift in this audit is the pole concept for variant 08 under the steering SUT (\num{0.0205}).
These hardening checks are descriptive safeguards for the N=96 exact-geometry subset (three Virtual KITTI scenes), not corpus-wide guarantees; the full tabular evidence is in the reproduction package.

\subsection{External validity on held-out scenes}
\label{sec:tiv_external_validity}
The main benchmark and the hardening analysis above are drawn from Virtual KITTI scenes \texttt{01}/\texttt{02}/\texttt{06}.
To check that the two central empirical patterns are not specific to those scenes, we re-evaluated all ten audited checkpoints on a disjoint held-out set, Virtual KITTI scenes \texttt{18} and \texttt{20} (60 exact-geometry pairs: 16 from scene \texttt{18}, 44 from scene \texttt{20}), using the identical exact-geometry DFF/C-DFF mechanism pipeline. These two scenes are held out from the ten generators' training but belong to the same Virtual KITTI corpus used in the authors' prior evaluation lineage, so this check bounds within-corpus generalization rather than transfer to an independent domain; a genuinely out-of-corpus paired dataset (e.g.\ a real--synthetic twin outside Virtual KITTI) remains future work.

First, the target-image parity pattern persists. The three steering-targeted checkpoints (variants \texttt{02}/\texttt{05}/\texttt{08}) remain the weakest reconstructions of the matched Virtual KITTI targets on the held-out scenes (mean target MAE \num{0.307}, mean SSIM \num{0.237}), while the baseline and YOLOP-targeted checkpoints stay close to their targets (mean target MAE \num{0.098}, mean SSIM \num{0.535}), i.e.\ the same ordering observed on the in-distribution benchmark.

Second, and more importantly for the diagnostic claim, output and appearance fidelity remain non-redundant with decisive-mechanism distance. On the held-out scenes, output consistency is essentially decoupled from DFF for every system under test: the per-SUT Spearman rank correlations between OV and DFF are $\rho=-0.072$ (steering), $-0.051$ (YOLOP--DA), and $-0.019$ (YOLOP--LL), and all three bootstrap $95\%$ CIs include zero ($n=600$ rows per SUT over the ten checkpoints; these intervals resample rows rather than pair clusters and are therefore anticonservatively narrow---since they nonetheless include zero, a correctly pair-clustered and hence wider interval would include zero as well). Perceptual SSIM-to-target is likewise uninformative about DFF for steering ($\rho=+0.052$) and YOLOP--LL ($\rho=-0.006$), with the only non-trivial appearance correlation being YOLOP--DA ($\rho=-0.405$). The mechanism magnitudes themselves stay small with dispersion comparable to their means (held-out DFF MSE $3.59\times10^{-4}{\pm}2.51\times10^{-4}$, $9.68\times10^{-5}{\pm}7.35\times10^{-5}$, $2.27\times10^{-5}{\pm}1.31\times10^{-5}$ for steering/YOLOP--DA/YOLOP--LL), so the population-level screening interpretation -- rather than fine-grained generator ranking -- carries over to scenes the generators never produced training targets for. The non-redundancy of DFF with both output- and appearance-level fidelity is therefore not an artifact of the three main scenes.

\subsection{Qualitative dissociation evidence}
\begin{figure*}[t]
\centering
\includegraphics[width=\textwidth]{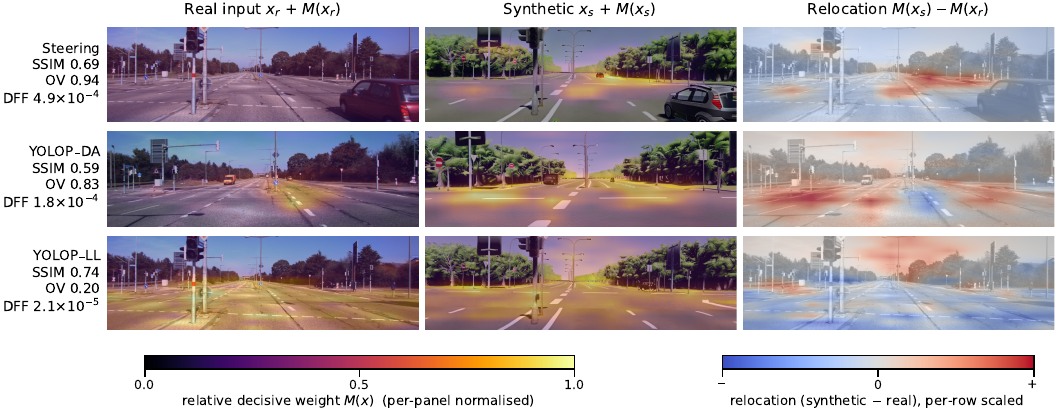}
\caption{Output similarity does not imply mechanism parity. Each row is an audited witness (one per frozen SUT: a baseline-generator ``high apparent fidelity / high mechanism mismatch'' case under the hardening rule). Output consistency (OV) is on a per-SUT scale---benchmark means are \num{0.32} (steering), \num{0.45} (YOLOP--DA), and \num{0.12} (YOLOP--LL)---so each witness sits at high apparent fidelity \emph{for its SUT} (OV \num{0.94}/\num{0.83}/\num{0.20} against those means). The real input $\xr$ and synthetic counterpart $\xs$ are close in image and per-SUT output terms, yet the frozen-SUT decisive maps $\Mmap(\xr)$, $\Mmap(\xs)$ (inferno overlays, columns~1--2) relocate, as the paired DFF and the signed difference $\Mmap(\xs)-\Mmap(\xr)$ (column~3) quantify. All panels use the audited benchmark decode and cached decisive maps.}
\label{fig:dissociation_grid}
\end{figure*}

To make the output--mechanism dissociation directly visible, \Cref{fig:dissociation_grid} shows three audited ``high apparent fidelity / high mechanism mismatch'' witnesses---one per frozen SUT, each a baseline-generator case that satisfies the pre-registered hardening rule of \Cref{sec:tiv_diagnostic_value}. The three rows are constrained to the baseline generator so that they share a single generator, and are selected by the rule rather than by visual appeal; they are illustrative rule-satisfying witnesses, not the single global top-ranked frame across all generators.
In every row the real input $\xr$ and its synthetic counterpart $\xs$ are close in image and output terms (annotated SSIM and output consistency), yet the frozen-SUT decisive maps $\Mmap(\xr)$ and $\Mmap(\xs)$ concentrate their weight in different locations; the signed difference in the third column is exactly the relocation the paired DFF value measures.
All panels use the audited benchmark decode and cached decisive maps, so the figure is traceable to the same evidence as \Cref{tab:final_retrain_benchmark,tab:final_mechanism_metrics}, and the featured witnesses are the strongest disagreement rows rather than visually flattering ones.

\begin{figure*}[t]
\centering
\includegraphics[width=\textwidth]{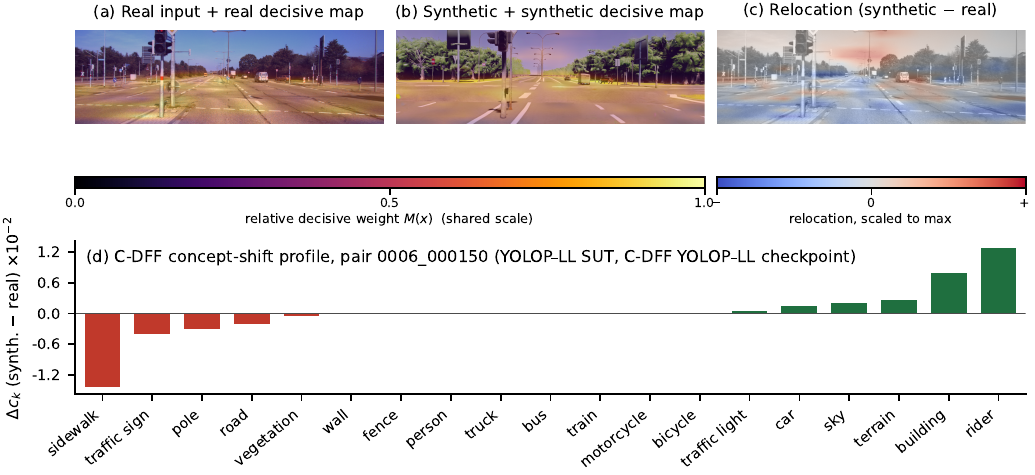}
\caption{Mechanism-level diagnostic for pair \texttt{0006\_000150} on a non-degraded generator (the C-DFF YOLOP--LL checkpoint). (a)~frozen YOLOP--LL decisive map on the real input; (b)~the same on the synthetic counterpart; (c)~signed difference (synthetic $-$ real); (d)~per-concept C-DFF importance shift. The illustrative single pair visualizes the output--mechanism dissociation; interpretation is given in the text above.}
\label{fig:decisive_concept}
\end{figure*}

\begin{figure}[t]
\centering
\includegraphics[width=\columnwidth]{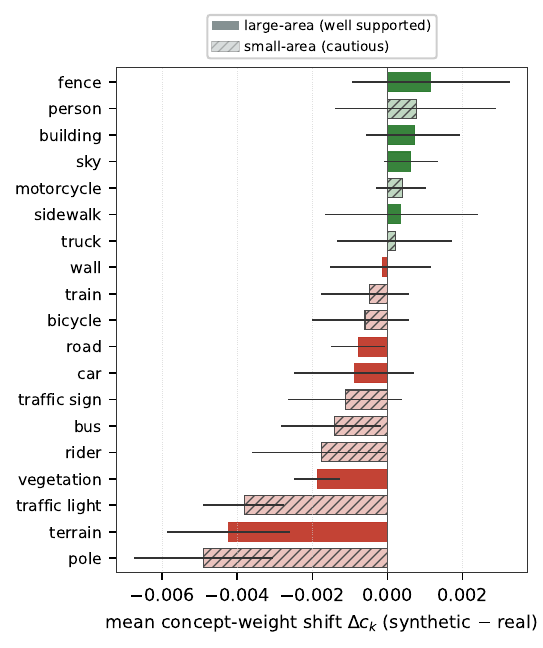}
\caption{Population-level per-concept C-DFF shift: mean concept-importance change $\Delta c_k$ (synthetic $-$ real) across the $96$ benchmark pairs, ranked, with pair-clustered bootstrap $95\%$ CIs. Roadside vertical cues lose decisive weight in the synthetic domain (red) while broad surfaces gain it (green). Small-area classes (hatched) are shown for completeness but interpreted cautiously per the $16\times16$ support caveat.}
\label{fig:concept_aggregate}
\end{figure}

To zoom into a single audited pair, \Cref{fig:decisive_concept} renders the decisive-evidence and concept-level diagnostics for a non-degraded generator (the C-DFF YOLOP--LL checkpoint, pair \texttt{0006\_000150}) rather than the RGB outputs alone.
Panels (a) and (b) overlay the frozen YOLOP--LL decisive maps on the real input and its synthetic counterpart; panel (c) shows their signed difference, so the relocation of decisive evidence is directly observable even where the two frames look broadly similar.
Panel (d) summarizes that same pair's C-DFF concept-importance shift, recomputed from the final exact-geometry cache: sidewalk, traffic sign, pole, and road lose decisive weight in the synthetic frame, while rider, building, terrain, and sky gain it (eight classes are unsupported on this pair's pooled grid and are exactly zero). These shifts are of order $10^{-2}$; \Cref{tab:final_mechanism_metrics} reports their mean square over the $K$ concepts, $\num{2.50e-5}$ here, so a $10^{-2}$ shift axis and a $10^{-5}$ C-DFF column describe the same evidence at different powers.
Because concept importance is aggregated on the $16\times16$ decisive-map grid, small-area classes (e.g.\ rider, person, traffic light) occupy only a few cells and their per-concept shifts are correspondingly noisier than those of large-area classes (road, building, sky, vegetation)---the largest single gain in panel (d), rider, is exactly such a low-support class; the per-concept reading is therefore qualitative, and quantitative weight should be given to the well-supported large-area concepts. Aggregated across the benchmark (\Cref{fig:concept_aggregate}), the population pattern is related but not identical: roadside vertical cues (pole, terrain, traffic light, vegetation) systematically lose decisive weight, with pair-clustered intervals excluding zero, whereas the positively shifting broad surfaces (sky, building, fence) have intervals spanning zero. Individual pairs depart in both directions---here terrain and traffic light move opposite to their population means---so panel (d) illustrates the diagnostic rather than evidencing the aggregate.


\subsection{Positive control and estimator sanity checks}
\label{sec:tiv_positive_control}
The dissociation results above are a decoupling---a null relationship between output and mechanism fidelity. On their own they do not show that the mechanism metric \emph{fires} when a mechanism genuinely shifts. We therefore add a constructed positive control with a known ground-truth mechanism, together with standard estimator sanity checks. All of these drive the same deployed counterfactual estimator used throughout the paper, not a differentiable surrogate.

\begin{figure*}[t]
\centering
\includegraphics[width=\textwidth]{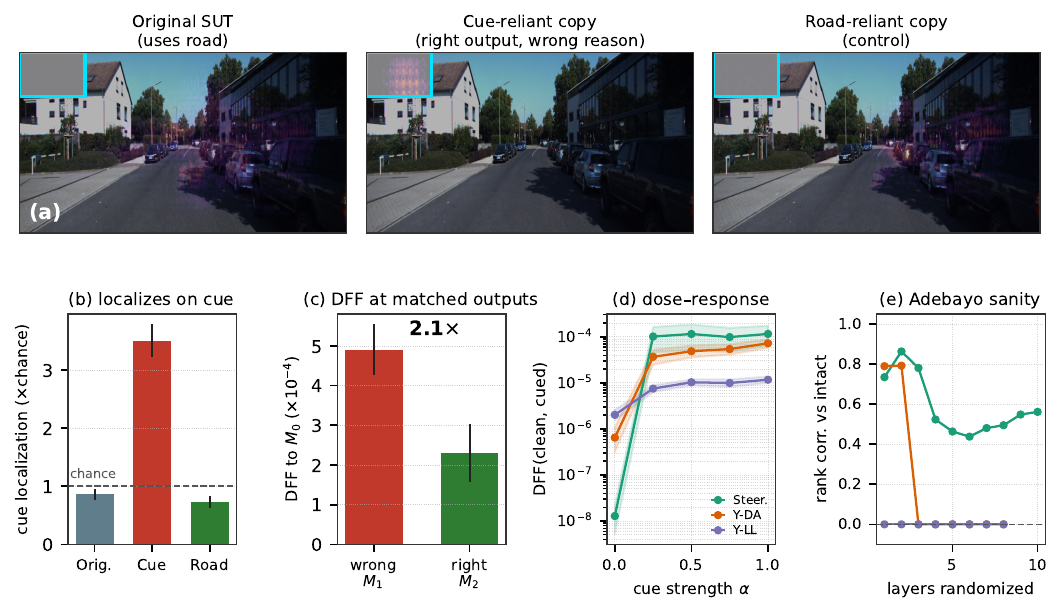}
\caption{Positive control and estimator sanity checks (steering SUT). (a)~On one held-out frame, with the planted cue set to encode the original SUT's own prediction so all models agree on the output, the deployed decisive maps of the original SUT and of a road-reliant copy fall on the road, whereas the cue-reliant copy's map concentrates inside the planted-cue region (cyan box). (b)~Cue localization (decisive-mass fraction inside the cue region divided by its area; $1\times$ is chance): $3.5\times$ for the cue-reliant copy versus $0.9\times$ (original SUT) and $0.7\times$ (road-reliant copy), both near chance (means over $60$ frames, $95\%$ CI whiskers). (c)~At matched output consistency (OV~$\num{0.66}$ for both copies), DFF to the original fires $2.1\times$ higher on the wrong-reason (cue) copy than on the right-reason (road) copy. (d)~Injecting a localized cue of increasing strength $\alpha$ raises DFF between the clean and cued decisive maps for all three SUTs. (e)~Adebayo model-parameter randomization: cascading weight randomization drives the steering decisive map's rank correlation with the intact-model map from $\num{1.00}$ to $\num{0.56}$ at full randomization (mid-cascade minimum $\num{0.44}$); the YOLOP-head maps become near-constant under randomization, so their rank correlation is undefined (${\approx}0$).}
\label{fig:positive_control}
\end{figure*}

\emph{Right output, wrong reason.} We fine-tune two copies of the steering SUT while keeping its architecture fixed. A cue-reliant copy is trained to read a planted spurious cue---a small patch in a fixed image region whose intensity encodes the steering label---with the road decorrelated from the label during training, so the cue is its only predictor; a road-reliant copy is trained on the same cued images but with the cue made uninformative, so it must keep using the road. Post-hoc diagnostics confirm the intended reliance: the cue-reliant copy tracks the encoded cue (prediction-to-cue correlation \num{0.99}) and is nearly invariant to the road (prediction standard deviation \num{0.011} as the road varies with the cue fixed), while the road-reliant copy tracks the original SUT across frames (output correlation \num{0.92}). On $60$ held-out frames we set the cue to encode the original SUT's own prediction, so all three models agree on the output---the original's output consistency with the cue-reliant and with the road-reliant copy is \num{0.66} in both cases. Under these matched outputs the deployed decisive map of the cue-reliant copy concentrates on the planted cue, localizing there at $3.5\times$ the chance mass ($95\%$ CI $[3.2,3.8]$), against $0.9\times$ (chance) for the original SUT and $0.7\times$ for the road-reliant copy, and exceeding twice the road-reliant copy's localization on every one of the $60$ frames; correspondingly, the DFF between the original and the cue-reliant copy is $2.1\times$ that between the original and the road-reliant copy ($\num{4.9e-4}$ versus $\num{2.3e-4}$). Because the planted cue occupies a fixed top-left region (\num{4.9}\% of the frame) that road-aligned evidence largely avoids, the road-reliant copy's below-chance localization partly reflects that placement; the cue-reliant copy's above-chance concentration on the cue and the paired DFF separation do not depend on it, and cue-location robustness (e.g.\ road-overlapping placements) is left to future work. The metric therefore fires selectively on a matched-output but mismatched-mechanism model that output agreement alone cannot distinguish, and its decisive map points at the true, planted mechanism. This is the positive-sensitivity evidence a purely dissociation-based argument lacks. Because a matched-output/wrong-reason copy is well posed only for the scalar steering output, this control establishes positive sensitivity directly for the steering SUT; for the dense YOLOP heads an analogous construction did not transfer (a documented negative result), so their positive sensitivity rests on the dose-response and randomization checks below and the population-level decoupling. \Cref{fig:positive_control} summarizes all of these controls.

\emph{Cue-injection dose-response.} As a complementary check we inject a localized cue of increasing strength into benchmark synthetic images and drive the same estimator. DFF between the clean and cued decisive maps rises monotonically with cue strength for all three SUTs, tracking the induced output shift; because any image perturbation moves DFF, this dose-response alone does not separate a task-relevant cue from a placebo, which is why the fine-tuned-copy control above---with its ground-truth localization---is the primary positive evidence.

\emph{Model-randomization sanity check.} Following \cite{Adebayo2018Sanity}, we cascadingly randomize the frozen SUT weights from the output layer inward and recompute the decisive map. Its rank correlation with the intact-model map falls from \num{1.00} to \num{0.56} at full randomization for steering, through a mid-cascade minimum of \num{0.44}; for the YOLOP heads the randomized maps become near-constant, so their rank correlation is undefined rather than a graded decorrelation. The decisive map thus depends substantially on the trained SUT weights, the residual correlation at full randomization reflecting the input-geometry sensitivity that gradient-based explanations are known to retain.

\emph{Relocation versus magnitude.} Because DFF is an unnormalized pooled-map distance, it mixes decisive-evidence \emph{relocation} with overall perturbation \emph{magnitude}. Decomposing it on the benchmark, DFF correlates with both the change in total map mass (Spearman \num{0.50}--\num{0.75} across SUTs) and a magnitude-free, $L_1$-normalized relocation distance (\num{0.13}--\num{0.86}), and the mix varies by SUT---YOLOP--DA is magnitude-dominated, the others relocation-dominated. Importantly, the output--mechanism decoupling is not a magnitude artifact: recomputing the OV-versus-mechanism correlations with the normalized relocation-only distance, or with a scale-invariant cosine distance, leaves them essentially unchanged (within \num{0.05} of the raw-DFF values for every SUT).

\emph{Pooling-grid ablation.} The $16\times16$ decisive-map grid is a fixed design choice, so we recompute full-resolution decisive maps on a $36$-pair subsample and re-pool at $h=w\in\{8,16,24,32\}$. The output--mechanism decoupling is grid-invariant: the OV-versus-DFF Spearman correlation varies by at most \num{0.04} across grids for every SUT (steering $\approx\!+0.23$, YOLOP--DA $\approx\!-0.20$, YOLOP--LL $\approx\!0$ on this subsample), and the per-generator DFF ordering is stable (Kendall $\tau\ge\num{0.86}$ against the primary $16\times16$ grid for all SUTs). Neither the decoupling nor the generator ordering is therefore an artifact of the pooling resolution.

\emph{Compute cost.} A deployed counterfactual decisive map takes about \num{1}~second per image per SUT on a single NVIDIA L40 GPU. A full mechanism-parity screen of the $96$-pair benchmark---one real-image map per pair, ten generated-image maps, across three SUTs---is on the order of \num{3000} counterfactual optimizations, roughly \num{50} minutes end to end, so the screen is inexpensive relative to collecting the matched real driving data it is intended to reduce reliance on.

\subsection{Reproducibility and result integrity}
\label{sec:model_specific_audit}
To keep the empirical scope auditable, the paper reports only checkpoint evidence traceable to the audited final output cache: every image, aggregate, and DFF/C-DFF value is recomputed from the final ten checkpoint directories, and no quantity is carried over from any external source.
Because several checkpoints share architecture, tensor count, and file size, a model-specific integrity audit verifies that all ten checkpoint weight hashes are unique, each generated-record directory matches its manifest model directory, every final benchmark PNG has a per-pair unique SHA-256 hash, and image, aggregate, and DFF/C-DFF metrics recompute from the cached images, per-image metrics, decisive maps, and segmentation tensors within \num{1e-5}, with no blocking mismatch.
A stricter provenance ledger additionally hashes the model-registry artifacts, all 960 final quantitative and 120 fresh qualitative PNG records, and \num{11520} mechanism-cache references, recording zero blocking issues.

\section{Discussion}
\label{sec:discussion}
DFF and C-DFF are not interchangeable metrics. DFF is closer to the raw mechanism signal because it directly compares decisive evidence maps; C-DFF operates on a derived representation---the same decisive evidence aggregated over semantic concepts---so it is less direct but interpretable in scene terms.

\begin{figure*}[t]
\centering
\includegraphics[width=\textwidth]{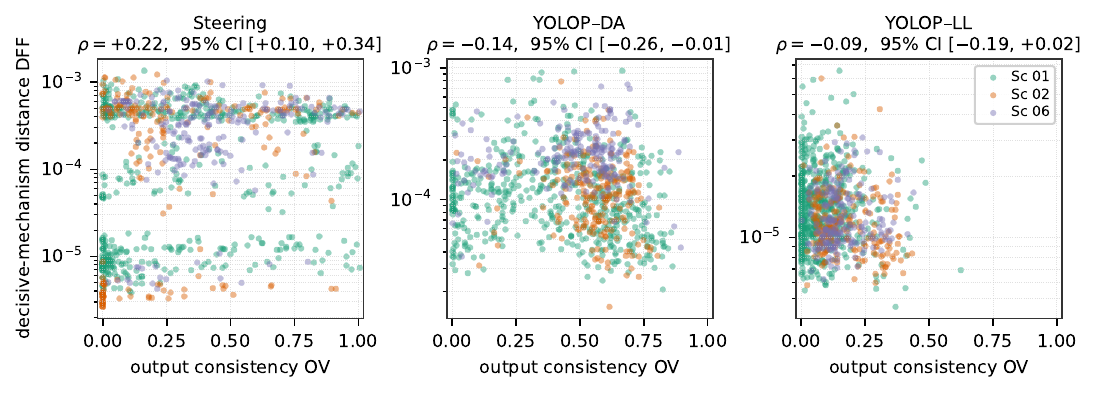}
\caption{Output--mechanism decoupling on the in-distribution \mbox{$N{=}96$} benchmark. Each point is one generator--pair evaluation ($960$ per SUT), colored by scene; output consistency OV (horizontal) is plotted against decisive-mechanism distance DFF (vertical, log scale). The pair-clustered Spearman $\rho$ with $95\%$ CI is annotated per SUT. The point clouds are essentially flat and mixed in sign ($\rho=+0.22$, $-0.14$, $-0.09$), so output agreement does not determine mechanism agreement.}
\label{fig:decoupling_scatter}
\end{figure*}

The disagreement audit makes the need for that second gate concrete. Using the hardening rule, 426 of the 2880 model--SUT evaluations ($14.8\%$) fall into a ``high apparent fidelity / high mechanism mismatch'' region: cases whose outputs look acceptable yet whose decisive evidence has moved. This fraction modestly but significantly exceeds the $13.5\%$ rate expected if apparent fidelity and mechanism mismatch were statistically independent (estimated by permuting the mechanism ranks against the apparent-fidelity ranks; $95\%$ null interval $[\num{0.126},\num{0.143}]$, permutation $p<\num{1e-3}$), so the two are weakly \emph{positively} associated rather than independent, though the excess over independence is small---a metric that simply tracked apparent fidelity would instead drive this fraction toward zero. Its practical value is that it isolates concrete witness cases that can be localized and inspected. The substantive dissociation evidence is the effect-size analysis that follows. On the in-distribution \mbox{$N{=}96$} benchmark, output consistency (OV) and decisive-mechanism distance are at most weakly rank-correlated for every system under test, and the sign is not even consistent across systems. Because the ten generators evaluated per paired scene are not independent observations, we resample the $96$ paired scenes as the clustered bootstrap unit; the resulting pair-clustered Spearman correlations are $\rho=+0.22$ for steering ($95\%$ CI $[+0.10,+0.34]$), $\rho=-0.14$ for YOLOP--DA ($[-0.26,-0.01]$), and $\rho=-0.09$ for YOLOP--LL ($[-0.19,+0.02]$), recomputed directly from the audited mechanism cache and visualized in \Cref{fig:decoupling_scatter}. We read these as descriptive effect sizes rather than attaching a per-SUT significance test, because a coherent single-unit test requires collapsing the ten non-independent generators to one OV/DFF value per pair ($n=96$): under that coherent pair-level test only the steering association is a clear positive effect ($\rho=+0.33$, $[+0.15,+0.49]$), while the YOLOP heads are weak and unit-dependent in significance ($\rho=-0.15$, $p=\num{0.15}$, for YOLOP--DA; $\rho=-0.21$, $[-0.39,-0.01]$, for YOLOP--LL): the pair-clustered and coherent units disagree on whether each YOLOP interval excludes zero, so we claim no per-SUT significance for the YOLOP heads. An earlier scene-stratified permutation applied to the pseudo-replicated $960$ rows overstated per-SUT significance and is not relied upon. The associations are small (squared rank correlations at most $\approx\!11\%$, i.e.\ little shared rank structure) and mixed in sign, so output agreement does not determine mechanism agreement. This dissociation is not manufactured by the three failed steering-targeted generators (variants \texttt{02}/\texttt{05}/\texttt{08}): excluding them leaves only the small steering association ($\rho=+0.29$, $[+0.12,+0.43]$) while the YOLOP--DA and YOLOP--LL correlations shrink toward zero with confidence intervals spanning zero ($\rho=-0.07$ and $\rho=-0.06$), so removing the broken generators weakens rather than creates the apparent coupling. On the disjoint held-out scenes \texttt{18}/\texttt{20} the association weakens further to statistically indistinguishable from zero (\Cref{sec:tiv_external_validity}), i.e.\ output fidelity is then uninformative about mechanism distance. This also holds against a learned perceptual baseline \cite{Zhang2018LPIPS}: LPIPS-to-target is at most weakly rank-correlated with DFF ($|\rho|\le 0.22$ per SUT), so mechanism distance is not recoverable from perceptual realism either. It further holds against a latent-feature baseline: the distance between the frozen SUT's own intermediate features on the real and synthetic inputs (cosine distance on the steering convolutional stack and on a mid-network YOLOP feature) is likewise at most weakly rank-correlated with DFF ($|\rho|\le 0.20$ per SUT over $n=960$ rows), so mechanism distance is not recoverable from broad internal-representation distance either. The dissociation witnesses of \Cref{fig:dissociation_grid} show this directly: a generated image can look close to its target and preserve the SUT output while redirecting the frozen SUT's decisive evidence.

In a development workflow, DFF/C-DFF can act as a mechanism-fidelity screen: apply conventional image/output checks first, then use DFF/C-DFF to flag scenes where outputs match but mechanisms diverge, and escalate high-disagreement cases for manual review; after any recalibration, the hardening checks should be rerun before claiming improvement. The aim is to improve validation \emph{efficiency} and \emph{interpretability}---reducing reliance on costly real data while retaining mechanism fidelity---not to provide a safety verdict; safety sufficiency requires a separate assessment that does not rest on (C-)DFF or other SUT-specific fidelity metrics.

The threshold-sensitivity audit also shows that the interpretation is not pinned to a single percentile cutoff. On the steering DFF screen, the three steering-targeted variants (OVF steering, DFF steering, and C-DFF steering concept) are the weakest at the 75th-percentile cutoff, with pass rates of \num{0.615}, \num{0.698}, and \num{0.688}, whereas the concept-guided C-DFF YOLOP--LL variant is the most stable, rising from \num{0.865} at the 75th percentile to \num{0.990} at the 95th; OVF steering is the most threshold-sensitive, reaching \num{0.917} at the 95th. This is useful operationally: stable pass patterns strengthen a validation claim, while threshold-sensitive variants should be treated as manual-review candidates rather than clean passes.
Operationally, the audit trail is part of the validation claim rather than an implementation detail: for each scene admitted to a mechanism-parity review, the evaluator should retain the pair provenance, checkpoint and SUT identifiers, decisive-map and segmentation parameters, exact crop/pad mapping, and per-pair DFF/C-DFF beside the conventional image and output metrics, so a disagreement case stays reproducible when any pipeline component changes. Conversely, a low DFF/C-DFF value is only the narrower statement that one frozen SUT exhibited closer decisive evidence on one declared paired benchmark, not evidence of safety transfer. The conservative decision rule is therefore asymmetric: DFF/C-DFF can block, prioritize, or explain simulation evidence when mechanisms disagree, but passing values only reduce one class of simulation-fidelity concern.

The present claims are restricted to paired monocular KITTI--Virtual KITTI~2 evaluation, three frozen SUTs, ten retrained ControlNet generators, and the declared 96-pair (three-scene) exact-geometry benchmark. The manuscript does not claim corpus-level generalization, multi-sensor validity, temporal validity, planning-level safety transfer, or sufficiency of low DFF/C-DFF for safety. Those extensions require new evidence on broader datasets and semantic taxonomies \cite{Caesar2020NuScenes,Sun2020Waymo,Yu2020BDD100K,Caesar2018COCOStuff,Zhou2017ADE20K}, and on safety-case integrations aligned with SOTIF- and ANSI/UL~4600-style assurance practice \cite{ISO21448_SOTIF,UL4600,Burton2020MissionSafety,Abrecht2024SafetyConcerns}.

\section{Threats to Validity}
\label{sec:threats}
DFF and C-DFF inherit approximation error from the explanation module: counterfactual maps depend on objective design, optimization horizon, regularization, random seeds, and target-side definitions. To bound the concern that a decisive map is an artifact of the single deterministic counterfactual initialization, we ran a restart-stability diagnostic re-running the search five times per frame from seeded Gaussian-perturbed start points on six scene-stratified KITTI frames. Pooled decisive maps are stable in shape and rank, with mean pairwise cosine/Spearman across restarts of \num{0.905}/\num{0.824} (steering), \num{0.960}/\num{0.918} (YOLOP--DA), and \num{0.968}/\num{0.928} (YOLOP--LL); steering is least stable, consistent with its weaker generators. Restart variability is thus a quantified, limited source of variation. To further test that the output/mechanism dissociation is not an artifact of the single counterfactual estimator, we recomputed the mechanism distance on the held-out scenes \texttt{18}/\texttt{20} with a second, independent explainer---vanilla input-gradient saliency of the frozen SUT---over the same cached images. The output-fidelity-to-mechanism Spearman correlation stays near zero to weakly negative under this alternative explainer (steering \num{-0.054}, 95\% CI $[-0.131,+0.024]$; YOLOP--DA \num{-0.117}, $[-0.196,-0.036]$; YOLOP--LL \num{-0.294}, $[-0.367,-0.219]$; $n=600$ per SUT, row-resampled intervals as in \Cref{sec:tiv_external_validity}), reproducing the decoupling found with the counterfactual estimator; because these row-resampled intervals are anticonservative, we do not read any of these per-SUT correlations as establishing a nonzero association; the robust statement is only that the sign of the output-to-mechanism relationship stays near zero to weakly negative under the alternative explainer, matching the counterfactual estimator. Cross-explainer concordance between the two mechanism estimates is, however, heterogeneous---strong for YOLOP--DA (Spearman \num{0.874}, $[0.844,0.897]$) but only weakly positive for steering (\num{0.128}, $[0.044,0.210]$) and YOLOP--LL (\num{0.109}, $[0.028,0.188]$)---so the qualitative decoupling is robust to explainer choice while the magnitude of the mechanism estimate is not, and DFF/C-DFF accordingly provide a population-level screening signal rather than a precise per-generator mechanism score. C-DFF also depends on the frozen segmenter taxonomy and quality; segmentation-distribution and concept-shift audits expose the cached class assignments but do not replace real-side semantic ground truth.

The \emph{positive} sensitivity of the metric---that the mechanism distance rises when the decisive mechanism genuinely shifts---is established by the fine-tuned-copy control of \Cref{sec:tiv_positive_control}, which drives the deployed counterfactual estimator (not a differentiable surrogate) and shows DFF firing selectively on a matched-output but mismatched-mechanism model. What remains open is the narrower \emph{generator-side} construction: a localized spurious cue injected into the synthetic input that a system could rely on instead of the genuine scene evidence. A preliminary probe of that kind, which planted shortcuts through a differentiable input-gradient surrogate, did not produce a significant DFF/C-DFF response---an outcome we attribute to the surrogate rather than to the metric, because the deployed estimator optimizes a different counterfactual objective. A synthetic-input planted-shortcut control that drives that same objective and cleanly separates a task-relevant cue from a placebo perturbation therefore remains future work; alongside the fine-tuned-copy control, the restart-stability and multi-explainer diagnostics quantify estimator robustness.

Statistical precision is likewise limited by the 96-pair benchmark size: the bootstrap audit reported above shows non-negligible relative 95\% confidence-interval widths for the mechanism metrics. The results therefore demonstrate the presence and localization of disagreement cases rather than narrow population-level estimates of mechanism-fidelity gaps. Scene specificity is partially mitigated by the held-out replication in \Cref{sec:tiv_external_validity}: re-evaluating all ten audited checkpoints on disjoint Virtual KITTI scenes \texttt{18}/\texttt{20} reproduces both the target-image parity ordering and the output/mechanism non-redundancy, but this remains a two-scene check and is not a corpus-wide generalization claim.

The concept-projection layer introduces an additional dependency on the frozen segmenter and label taxonomy (quantified in the concept-shift audit). The largest shift recorded there is small in absolute terms but still enough to matter when two variants are otherwise close, so C-DFF conclusions should be interpreted together with the concept-shift and segmentation-distribution ledgers, especially for near-ties.

Pairing quality is another structural threat: if real and synthetic samples are not truly matched, pairwise DFF/C-DFF can conflate domain mismatch with correspondence error. Finally, low DFF/C-DFF does not itself imply end-to-end safety transfer because temporal coupling, controller behavior, long-horizon dynamics, and external conditions remain outside the current paired monocular benchmark \cite{Hendrycks2019Robustness}.

\section{Conclusions}
\label{sec:conclusion}
This paper presented a unified treatment of mechanism parity for simulation-based validation of autonomous-driving perception. DFF directly compares decisive evidence maps, while C-DFF projects decisive evidence into semantic concept space. The final empirical path is restricted to audited final checkpoints, exact-size generated outputs, regenerated DFF/C-DFF mechanism metrics, fresh screened qualitative panels, and model-specific provenance ledgers. Three observations ground the conclusion: disagreement cases in which high apparent fidelity accompanies high mechanism mismatch occur at a rate that modestly but significantly exceeds output--mechanism independence (426 of 2880, permutation $p<\num{1e-3}$); output--mechanism rank correlations are small and sign-inconsistent in-distribution and on held-out scenes, and mechanism distance is not recoverable from appearance or latent-feature similarity; and a constructed positive control shows the decisive-map estimator localizes on a known planted mechanism and separates a matched-output but mismatched-mechanism model from a matched-mechanism one. For practitioners, this positions decisive-evidence comparison as a low-cost screening layer for existing simulation-validation pipelines: it surfaces output-matched scenes whose decisive evidence has nonetheless relocated---cases that appearance and output checks pass---so that scarce real-world validation effort can be concentrated where mechanism fidelity is in doubt, while a low value narrows but does not certify simulation-fidelity risk. Within that paired monocular benchmark, the result is a clear validation principle: beyond visual realism and output agreement, simulation evidence should also test mechanistic realism.

\section*{Acknowledgment}
This work was supported by an EPSRC-funded studentship and Siemens Industry Software Netherlands BV, administered through WMG, University of Warwick. For the purpose of open access, the author(s) has applied a Creative Commons Attribution (CC BY) licence to any Author Accepted Manuscript version arising. AI-assisted tools were used for language editing, code/package auditing, repository-documentation checks, figure-drafting support, and manuscript revision; all scientific content was reviewed and approved by the authors, and no AI system was treated as an author.

\section*{Code and Data Availability}
Reproduction materials---the retrained checkpoints, benchmark outputs, and audit ledgers---will be deposited in a public archival repository with a citable DOI upon acceptance, together with environment details, hashes, scripts, manifests, and reproduction instructions; third-party datasets and model checkpoints are not redistributed unless their licenses explicitly permit it.

\section*{Conflict of Interest}
The authors declare no conflict of interest. All affiliations and funding are disclosed in this article, and the sponsor did not influence the study design, data analysis, interpretation of results, or the decision to submit this article.

\section*{Author Biographies}
\footnotesize
\noindent\textbf{Danial Safaei} received the M.Sc. degree and is a Ph.D. student at WMG, University of Warwick, Coventry, U.K. His research interests include safe and responsible AI, autonomous-vehicle safety, simulation-based perception validation, synthetic-data fidelity, and mechanism-parity diagnostics.\par
\noindent\textbf{Siddartha Khastgir} is Professor and Head of Safe Autonomy at WMG, University of Warwick, Coventry, U.K., with research on AI safety and safety of autonomous transport across land, aviation, and marine domains.\par
\noindent\textbf{Matthew D. Higgins} (Senior Member, IEEE) is a Professor at the University of Warwick, Coventry, U.K., and leads WMG's Connectivity and Communications Technology Research Group.\par
\noindent\textbf{Jeroen Ploeg} received the Ph.D. degree in mechanical engineering from Eindhoven University of Technology in 2014, with work on control of vehicle platoons, and is with 2getthere, Utrecht, The Netherlands.\par
\noindent\textbf{Mohsen Alirezaei} received the Ph.D. degree in 2011 and is an Associate Professor at Eindhoven University of Technology (TU/e), Eindhoven, The Netherlands, working on verification and validation of automated and cooperative automated vehicles.\par
\noindent\textbf{Son Tong} (Member, IEEE) received the Ph.D. degree from KU Leuven, Belgium, in 2012, and is a Senior Research Engineer and Project Manager at Siemens Digital Industries Software, Leuven, Belgium.\par
\noindent\textbf{Paul A. Jennings} is a Professor at WMG, University of Warwick, Coventry, U.K. His research spans connected and automated vehicles, scenario-based safety testing and verification and validation of automated driving systems, and experiential engineering.\par
\noindent\textbf{Xingyu Zhao} is an Associate Professor in Safety-Critical AI Systems at WMG, University of Warwick, Coventry, U.K., working on trust, verification and validation, testing, and safe/explainable AI for safety-critical systems.\par
\normalsize

\end{document}